\documentclass[sigconf]{acmart}

\AtBeginDocument{%
  }

\usepackage{graphicx}
\usepackage{amsmath,mathcomp}
\usepackage{comment}

\usepackage[ruled,vlined,linesnumbered,noend]{algorithm2e}

\newcommand{\textproc}[1]{\textsc{\selectfont #1}}

\newcommand{\Torimi}[1]{\textcolor{magenta}{[Torimi: #1]}}

\title{BasketLiDAR: The First LiDAR-Camera Multimodal Dataset\\ for Professional Basketball MOT}

\author{Ryunosuke Hayashi}
\affiliation{
  \institution{Keio University}
  \city{Yokohama}
  \country{Japan}
}
\email{hayashi.ryu430@keio.jp}

\author{Kohei Torimi}
\affiliation{
  \institution{Keio University}
  \city{Yokohama}
  \country{Japan}
}
\email{silver2bd@keio.jp}

\author{Rokuto Nagata}
\affiliation{
  \institution{Keio University}
  \city{Yokohama}
  \country{Japan}
}
\email{nagatarokuto@keio.jp}

\author{Kazuma Ikeda}
\affiliation{
  \institution{Keio University}
  \city{Yokohama}
  \country{Japan}  
}
\email{kazu2080@keio.jp}

\author{Ozora Sako}
\affiliation{
  \institution{Keio University}
  \city{Yokohama}
  \country{Japan}
}
\email{sako.ozora@keio.jp}

\author{Taichi Nakamura}
\affiliation{
  \institution{AISIN CORPORATION}
  \city{Kariya}
  \country{Japan}
}
\email{taichi.nakamura@aisin.co.jp}

\author{Masaki Tani}
\affiliation{
  \institution{AISIN CORPORATION}
  \city{Kariya}
  \country{Japan}
}
\email{masaki.tani@aisin.co.jp}

\author{Yoshimitsu Aoki}
\affiliation{
  \institution{Keio University}
  \city{Yokohama}
  \country{Japan}
}
\email{aoki@keio.jp}

\author{Kentaro Yoshioka}
\affiliation{
  \institution{Keio University}
  \city{Yokohama}
  \country{Japan}
}
\email{kyoshioka47@keio.jp}

\date{\today}

\renewcommand{\shortauthors}{Ryunosuke Hayashi et al.}
\renewcommand\footnotetextcopyrightpermission[1]{} 

\begin{document}

\begin{teaserfigure}
  \centering
  \includegraphics[width=\columnwidth]{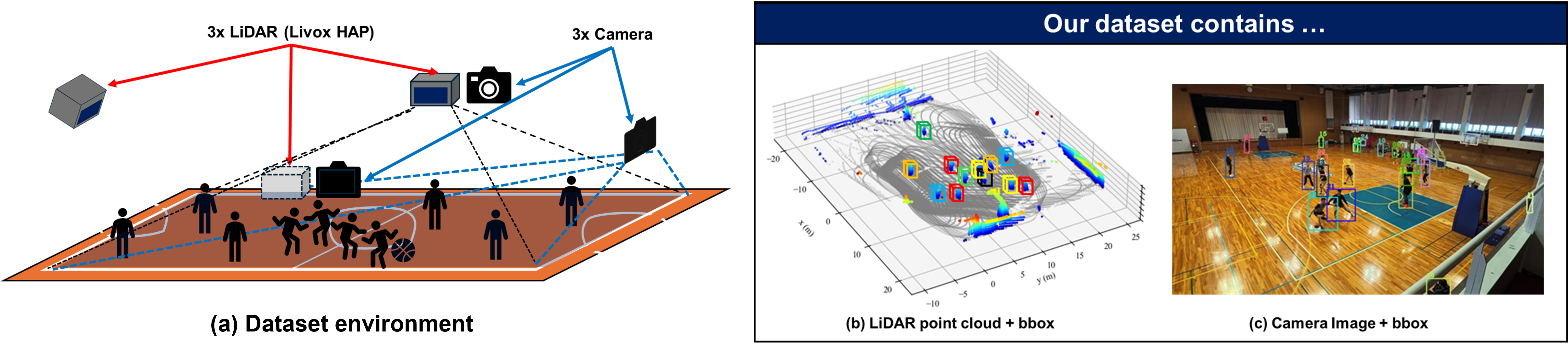}
  \caption{(a) BasketLiDAR captures comprehensive court coverage without blind spots using three sets of calibrated camera-LiDAR sensor systems. (b) BasketLiDAR data: LiDAR point clouds, multi-view camera images, and player bounding boxes in each point cloud with fully synchronized IDs across all modalities.}
\end{teaserfigure}


\begin{abstract}
Real-time 3D trajectory player tracking in sports plays a crucial role in tactical analysis, performance evaluation, and enhancing spectator experience. Traditional systems rely on multi-camera setups, but are constrained by the inherently two-dimensional nature of video data and the need for complex 3D reconstruction processing, making real-time analysis challenging. Basketball, in particular, represents one of the most difficult scenarios in the MOT field, as ten players move rapidly and complexly within a confined court space, with frequent occlusions caused by intense physical contact.

To address these challenges, this paper constructs BasketLiDAR, the first multimodal dataset in the sports MOT field that combines LiDAR point clouds with synchronized multi-view camera footage in a professional basketball environment, and proposes a novel MOT framework that simultaneously achieves improved tracking accuracy and reduced computational cost.
The BasketLiDAR dataset contains a total of 4,445 frames and 3,105 player IDs, with fully synchronized IDs between three LiDAR sensors and three multi-view cameras. We recorded 5-on-5 and 3-on-3 game data from actual professional basketball players, providing complete 3D positional information and ID annotations for each player. Based on this dataset, we developed a novel MOT algorithm that leverages LiDAR's high-precision 3D spatial information. The proposed method consists of a real-time tracking pipeline using LiDAR alone and a multimodal tracking pipeline that fuses LiDAR and camera data. Experimental results demonstrate that our approach achieves real-time operation, which was difficult with conventional camera-only methods, while achieving superior tracking performance even under occlusion conditions. \textbf{The dataset is available upon request at:} \href{https://sites.google.com/keio.jp/keio-csg/projects/basket-lidar}{\textit{https://sites.google.com/keio.jp/keio-csg/projects/basket-lidar}}
\end{abstract}


\maketitle

\section{Introduction}
Player trajectory tracking in sports analysis plays a central role in tactical analysis, performance evaluation, and enhancing spectator experience. Traditionally, coaches and analysts have relied on manual video analysis, but recent advances in computer vision technology have brought attention to automated approaches using Multiple Object Tracking (MOT). In particular, MOT, which performs player position estimation and ID association over time, has become an essential technological foundation for real-time tactical analysis, detailed player statistics generation, and immersive spectator experiences utilizing AR technology.

Traditional sports 3D MOT systems have primarily adopted multi-view camera approaches~\cite{wu2020multi}. These systems have employed MOT integrating multiple camera feeds to obtain accurate player position information and movement patterns necessary for tactical analysis.
However, since camera footage is inherently two-dimensional information, acquiring the 3D spatial information necessary for sports analysis requires multi-stage processing including person detection in each camera view, feature extraction, inter-camera correspondence, and 3D reconstruction. This complex processing pipeline incurs significant computational costs, creating major constraints for deployment in sports environments that require real-time performance. This represents a critical challenge that hinders the proliferation of data-driven sports analysis and value delivery to players, coaches, and spectators.

Therefore, this study poses the following research question:
\textbf{Can introducing LiDAR sensors simultaneously achieve improved accuracy and reduced computational cost in sports MOT?}

LiDAR sensors are 3D sensors with proven track records in person and vehicle detection in autonomous driving, possessing promising characteristics for solving sports MOT challenges. First, LiDAR can directly acquire high-precision 3D point clouds of surrounding environments at 10 FPS, eliminating the complex processing required to estimate 3D space from 2D information like camera footage. Second, since point cloud data itself has spatial structure, multi-sensor data integration can be achieved through extremely simple processing, eliminating the need for complex 3D reconstruction processing, likely enabling reduced computational cost for the entire system and real-time operation.

To validate this hypothesis, we selected basketball as our subject, where occlusions occur frequently and 3D analysis is particularly challenging. Basketball represents one of the most difficult scenarios in sports MOT, as ten players move rapidly and complexly within a confined court (28m×15m), with frequent occlusions caused by intense physical contact.

This paper makes the following contributions:
\begin{enumerate}
    \item \textbf{LiDAR-Camera multimodal sports MOT dataset:} We construct \textit{BasketLiDAR}, the first synchronized MOT dataset combining LiDAR point clouds and multi-view camera footage targeting professional basketball players, with ID and 3D position annotations for all players.
    \item \textbf{Proposal of fusion-based MOT framework:} We construct an multimodal MOT framework integrating LiDAR-based tracking with ID re-identification from camera footage, achieving real-time and high-precision MOT performance.
    \item \textbf{Research infrastructure for sports MOT:} We provide open-source publication of the dataset and the model implementation, contributing to research advancement in the LiDAR-based sports analysis field.
\end{enumerate}


\section{Related Works}

\subsection{Sports MOT Datasets}
Current sports MOT datasets are predominantly based on RGB videos. Representative examples include SportsMOT~\cite{cui2023sportsmot}, which is an MOT benchmark that annotates bounding boxes and IDs for all players on 240 monocular RGB videos (approximately 150,000 frames) containing indoor side-view footage similar to NBA broadcast angles. TeamTrack~\cite{scott2024teamtrack} is a large-scale benchmark using 4K-8K high-resolution fixed camera footage captured from diverse viewpoints including fisheye side views and drone overhead perspectives. TrackID3x3~\cite{yamada2025trackid3x3} is a dataset for integrated tracking and pose estimation evaluation that provides annotations including bounding boxes, IDs, and 10-joint 2D skeletal points for fixed camera and drone footage.

In the autonomous driving field, numerous multimodal datasets combining LiDAR and cameras have been constructed, including KITTI~\cite{geiger2012kitti}, Waymo~\cite{sun2020waymo}, and nuScenes~\cite{caesar2020nuscenes}. However, fundamental differences exist between autonomous driving and sports scenes in terms of tracking targets (vehicles vs. humans), motion patterns (predictable traffic rules vs. complex and unpredictable sports movements), and sensor placement (vehicle-mounted vs. fixed installation). Therefore, direct application of autonomous driving datasets to sports MOT is challenging, making the construction of specialized datasets addressing sports-specific challenges essential.

To summarize, all existing sports MOT datasets rely solely on RGB footage, with 3D positional information limited to indirect estimation from 2D imagery. In contrast, BasketLiDAR constructed in this study is the first multimodal dataset in the sports 3D MOT field with fully synchronized LiDAR point clouds and multi-view camera footage. Through direct 3D spatial measurement by LiDAR, it provides a novel research foundation enabling high-precision and real-time 3D MOT.

\subsection{Camera-based 3D MOT Technology}
Most MOT approaches for single cameras are implemented using Tracking-by-Detection (TBD), which has evolved by associating frame-by-frame detection results from object detectors with short-term state prediction using motion models and appearance information using Re-ID features~\cite{han2019re,he2023fastreid}. SORT and DeepSORT~\cite{bewley2016simple, wojke2017simple} introduced appearance-based similarity using Re-ID features into the SORT framework, suppressing ID switches in occlusion and high-density scenes. Recently, ByteTrack~\cite{zhang2022bytetrack} proposed a novel algorithm that associates previously discarded low-confidence detection results with existing tracks, significantly improving true positives for not losing targets under occlusion.

However, extending these single camera-based methods to multi-camera 3D MOT requires spatially aligning 2D tracking results obtained from each camera and integrating object position estimation and IDs in 3D space~\cite{dong2019fast}. The sports multi-camera 3D MOT framework proposed in Ref.~\cite{wu2020multi} converts 2D detection and tracking results from multiple cameras to 3D positions through triangulation, and fuses them to estimate 3D positions. However, this 3D cross-view integration presents significant computational challenges. Cycle consistency, which maintains consistency in both temporal and inter-camera directions simultaneously, requires complex optimization. More importantly, the triangulation process that reconstructs 3D positions from 2D detection results becomes a bottleneck occupying the majority of computation time (86\% of processing time in our implementation).

This study aims to eliminate the 3D position reconstruction process through direct 3D data acquisition by LiDAR, achieving low computational load and high-precision 3D position estimation to achieve real-time performance in sports 3D MOT.

\subsection{Multimodal Fusion Methods}
Multimodal fusion of LiDAR and cameras has been primarily studied in the autonomous driving field in the context of 3D object detection~\cite{chen2017multi, qi2018frustum} and tracking~\cite{weng2020ab3dmot}. These methods achieve high-precision object recognition that is difficult with single modalities by integrating LiDAR's high-precision 3D spatial information with cameras' rich appearance information.

However, direct application of these methods to sports scenes is challenging. Compared to autonomous driving scenes, sports scenes have unique challenges such as tracking targets being humans with fast and complex motion patterns, and frequent occlusions in dense environments. This study designs a novel multimodal MOT framework specifically addressing these sports scene-specific challenges.

\section{BasketLiDAR Dataset}

In this paper, we develop BasketLiDAR, a multimodal basketball MOT dataset combining LiDAR and cameras. This dataset simultaneously captures synchronized multi-view camera footage from three viewpoints and high-precision LiDAR point cloud data, targeting complex play scenes by professional basketball players. By providing player position information and ID annotations synchronized for all camera viewpoints and point cloud data, we offer a dataset that is differentiated from existing sports MOT datasets.

Table~\ref{tab:dataset_stats} shows the dataset statistics. BasketLiDAR constitutes a large-scale dataset with 4,445 total frames and 3,105 total IDs. The dataset is characterized by segmenting each play into short sequences (average 13.1 seconds) and targeting high-difficulty scenes that include frequent player occlusions and motion blur occurring in actual competitive environments. Such a dataset composed of multi-view camera footage and temporally synchronized LiDAR point clouds provides essential data for developing LiDAR-based sports tracking technology.

\begin{table}[t]
\centering
\caption{BasketLiDAR Dataset Statistics}
\label{tab:dataset_stats}
\begin{tabular}{lr}
\toprule
Metric & Value \\
\midrule
Total frames & 4,445 \\
Total bboxes & 397,757 \\
Total IDs & 3,105 \\
Sequences & 34 \\
Avg. sequence length & 13.1s \\
\bottomrule
\end{tabular}
\end{table}

\begin{figure}
  \centering
  \includegraphics[width=\columnwidth]{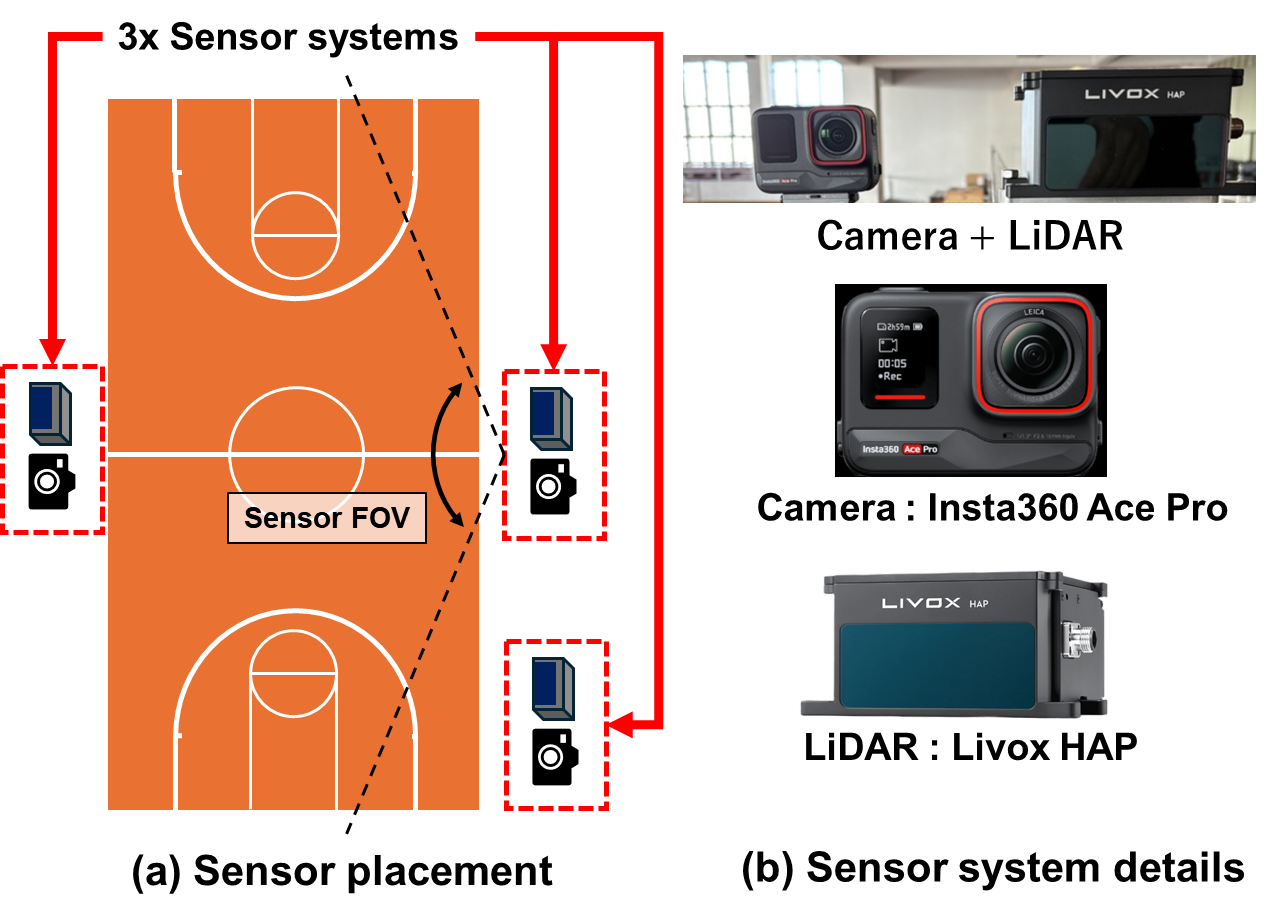}
  \caption{BasketLiDAR dataset recording environment. (a) Three sensor systems combining camera and LiDAR are positioned at different locations within the court. This arrangement is designed to provide comprehensive court coverage without blind spots, considering FOV constraints. (b) Sensor system details. Camera (Insta360 Ace Pro) and LiDAR (Livox HAP) are positioned adjacently.}
  \label{fig:setup}
\end{figure}

\subsection{Sensor setup}
As shown in Fig.~\ref{fig:setup}, we deployed three camera-LiDAR multi-sensor units to capture the entire basketball court. Each unit was installed to overlook the court from different directions, designed to minimize occlusions between players. In each unit, we fixed a camera (Insta360 Ace Pro~\cite{instaace}) and LiDAR (Livox HAP~\cite{livox-hap}) side by side and mounted them on tripods to ensure a stable data acquisition environment.

The capture specifications are as follows:
\begin{itemize}
\item \textbf{Frame rate}: 10 FPS for both camera and LiDAR (hardware synchronized)
\item \textbf{Camera specifications}: Resolution 3840×2160 (4K), FOV $95\tcdegree \times 78\tcdegree$, aperture F2.6
\item \textbf{LiDAR specifications}: Angular resolution $0.18\tcdegree \times 0.23\tcdegree$, 144 lines, point density 452,000 pts/s, FOV $125\tcdegree \times 25\tcdegree$, effective range 150m
\item \textbf{Calibration}: External parameters between three units pre-acquired
\end{itemize}

\subsection{Participants}
More than ten first-team players from SeaHorses Mikawa, belonging to Japan's professional basketball league B.League B1, participated in the dataset recording. These players cover all positions including Point Guard (PG), Shooting Guard (SG), Small Forward (SF), Power Forward (PF), and Center (C).

\subsection{Capturing Conditions}
In this study, we continuously filmed the practice sessions of professional basketball team SeaHorses Mikawa over two days. The filming targets included game formats such as 3-on-3 and 5-on-5, and repetitive set plays in front of the goal, incorporating menus with exercise intensity close to actual games and frequent player occlusions to collect natural and practical data. After filming, we manually examined these practice scenes and segmented them into short video clips (traces) averaging approximately 13.1 seconds per play.

\subsection{Annotation}
Annotation was performed on the collected camera frames from three units and LiDAR 3D point clouds according to the following guidelines:
\begin{itemize}
\item \textbf{Camera footage}: Individual annotation of bounding boxes and IDs for each of the three viewpoints
\item \textbf{LiDAR point clouds}: Unified annotation after integrating point clouds acquired from three LiDAR units in world coordinate system
\item \textbf{Target scope}: Players' limbs and entire torso as annotation targets, excluding other objects such as balls in contact with players
\item \textbf{Occlusion handling}: Predictive annotation of player bounding boxes even under occlusion, as long as part of the player's body is visible
\item \textbf{ID consistency}: Assignment of unique IDs to each player throughout the entire clip, ensuring consistency among annotators
\item \textbf{Annotation quality}: We achieved high-quality dataset construction by implementing multiple rounds of quality checks and corrections for all annotation results
\end{itemize}


\section{Proposed Method}
This section presents an overview of our proposed LiDAR-camera multimodal MOT framework. Our method primarily leverages LiDAR Bird's Eye View (BEV) information, which excels in spatial recognition, while maintaining compatibility with existing tracking-by-detection 2D trackers such as ByteTrack and CC-Sort. 

However, LiDAR alone cannot utilize appearance features, which limits identification performance for occluded targets. Therefore, our LiDAR-camera-fusion framework employs a two-stage approach that generates tracks rapidly using the LiDAR-only pipeline, then extracts only occlusion sessions and applies camera-based ID re-identification modules, thereby achieving both high speed and high ID consistency.

Section 4.1 describes the processing details of LiDAR-only tracking, Section 4.2 provides an overview of the LiDAR-camera-fusion tracking pipeline, and Sections 4.3 provide detailed explanations of each sub-module comprising the pipeline.

\begin{figure}[htbp]
  \centering
  \includegraphics[width=\columnwidth]{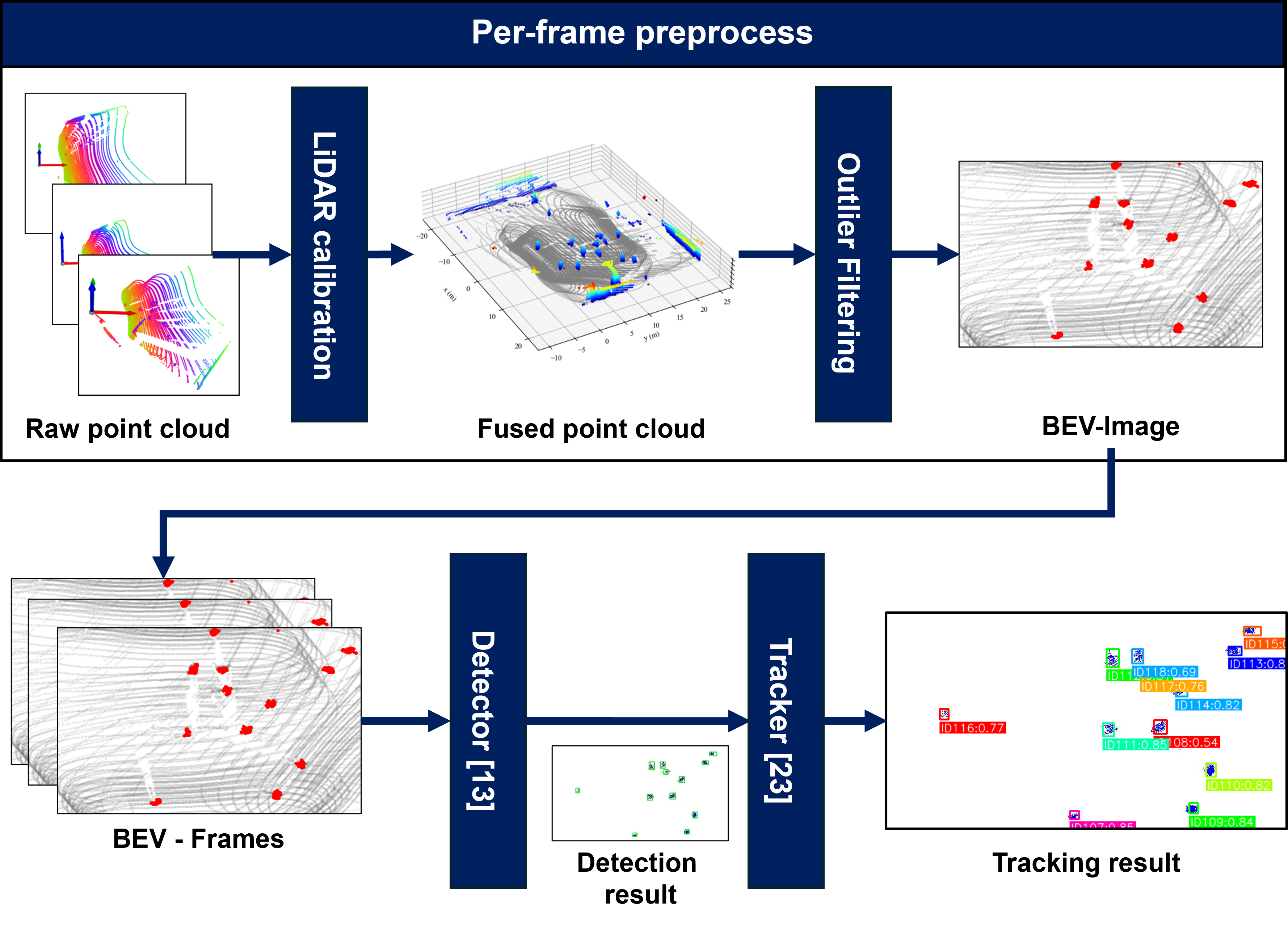}
  \caption{LiDAR-only MOT pipeline: Point clouds acquired from three LiDARs are integrated in a world coordinate system using calibration parameters, and only player regions are extracted through court region cropping and height filtering. Finally, BEV projection enables player detection using 2D detectors and ID association by trackers.}
  \label{fig:lidar-only}
\end{figure}

\begin{figure*}[!t]
  \centering
  \includegraphics[width=\textwidth,keepaspectratio]{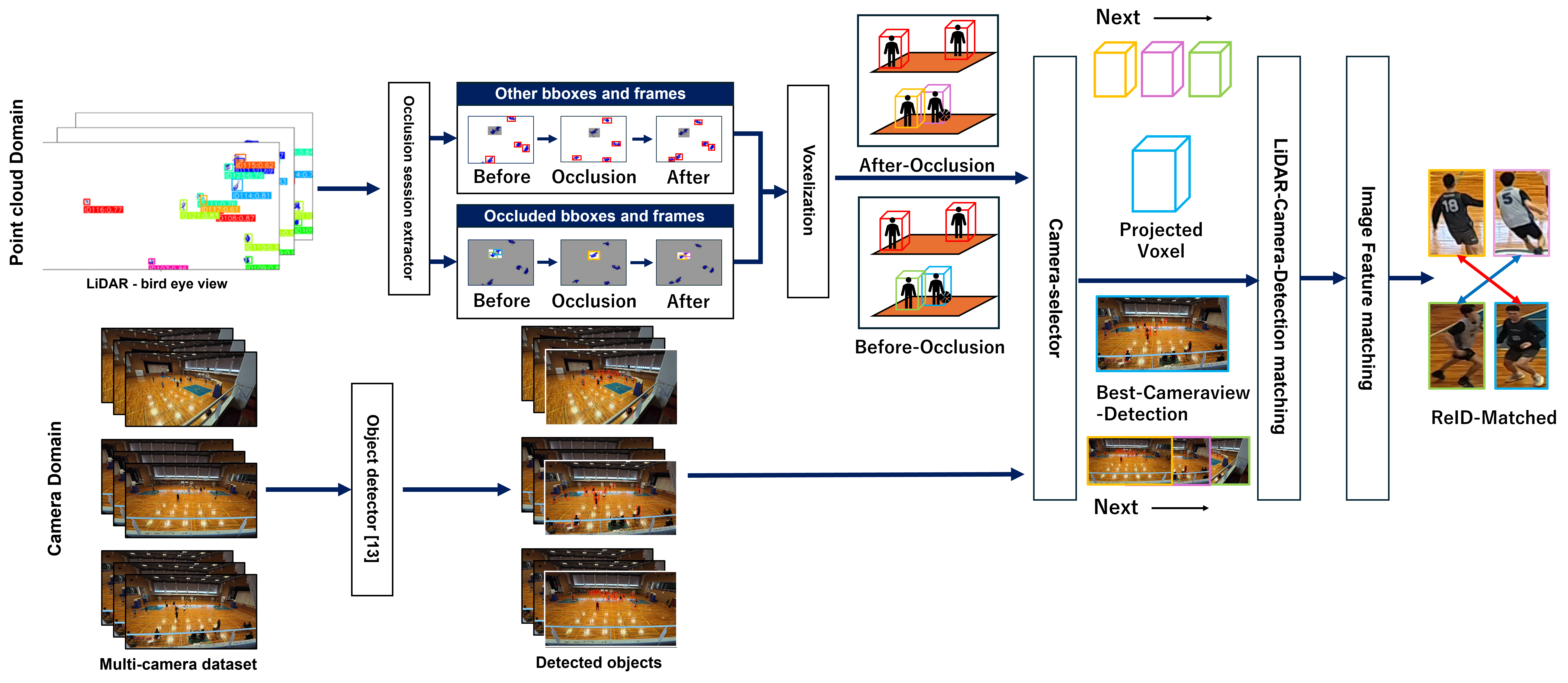}
  \caption{LiDAR-camera-fusion pipeline overview: Real-time tracking using LiDAR point clouds adaptively integrated with camera-based Re-ID for occlusion recovery. The Occlusion Session Extractor triggers appearance feature matching to restore correct ID associations when player occlusions occur.}
  \label{fig:lidar-fusion}
\end{figure*}

\subsection{LiDAR-only MOT pipeline}
Fig. \ref{fig:lidar-only} shows the LiDAR-only MOT pipeline. The pipeline consists of four stages: multi-LiDAR data integration, region filtering, BEV projection, and object detection and tracking.

\noindent \textbf{Multi-LiDAR Integration:} Point clouds $\mathbf{P}_i(t)$ ($i=1,2,3$) obtained from three LiDARs are integrated into a world coordinate system using pre-acquired extrinsic parameters (rotation matrix $\mathbf{R}_i$, translation vector $\mathbf{t}_i$).

\noindent \textbf{Region Filtering:} Unnecessary point clouds cause false detections, so we reduce false detections by removing point clouds that clearly do not constitute players. Specifically, we extract only players within the basketball court by removing floor and goalpost point clouds through height-direction filtering and excluding point clouds outside the court region in the XY plane.

\noindent \textbf{BEV Projection:} The filtered point clouds are projected onto a BEV map. While 3D object detectors that directly input point clouds exist~\cite{lang2019pointpillars}, they are designed for autonomous driving and have difficulty ensuring accuracy in player detection. Therefore, this study adopts a BEV projection approach to leverage 2D object detectors and trackers that have high versatility and generalize well with limited data.

\noindent \textbf{Object Detection and Tracking:} Player detection is performed on the BEV map using YOLOv11~\cite{yolov11} trained on BEV-projected player images from the BasketLiDAR dataset. The detection results are input to ByteTrack~\cite{zhang2022bytetrack} for temporal ID association.

\subsection{LiDAR-camera-fusion}
The point clouds of LiDAR provide an inexpensive means of capturing the 3D geometry information of the court, but they contain no appearance clues.
Consequently, LiDAR-only tracking frequently suffers ID switches, losses, or updates whenever player–player occlusion occurs.
To remedy this, we introduce a LiDAR–camera-fusion re-identification framework that leverages camera data before and after occlusion, as outlined in Figure~\ref{fig:lidar-fusion}.
The framework comprises three stages:
\begin{enumerate}
    \item Occlusion-Session-Extractor, which detects intervals of player occlusion.
    \item LiDAR–camera detection matching, which retrieves the frame indices in which each player is clearly visible immediately before and after the occlusion.
    \item RGB-based re-identification, which links the pre- and post-occlusion detections of the same player.
\end{enumerate}
The following subsections provide a detailed description of each stage.



\begin{figure*}[!t]
  \centering
  \includegraphics[width=\textwidth,keepaspectratio]{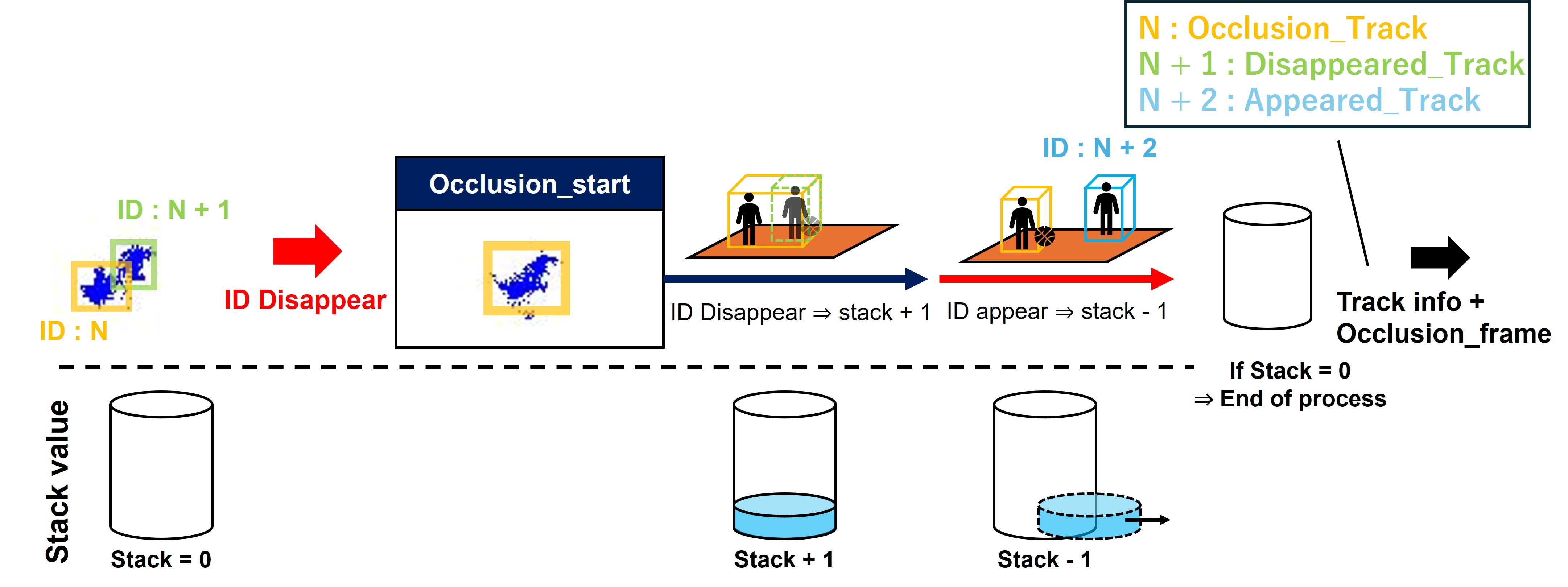}
  \caption{Occlusion-session-extractor: The module monitors the number of active track IDs frame by frame, opening an occlusion session whenever that count drops below its previous level and closing it once the count returns. For each session, it outputs the time span and the list of IDs involved in occlusion sessions together with their trajectories.
}
  \label{fig:fullwidth}
\end{figure*}

\subsubsection{Occlusion-Session-Extractor}
\label{sec:occlusion_session_extractor}
Let the size of the track-ID set at frame $t$ be
\begin{equation}
N_t := |\mathcal{T}_t|, \qquad t = 1,\dots,T
\end{equation}
Occlusion intervals are extracted from this time series.
An occlusion interval $I_k = [t_s^{(k)},\,t_e^{(k)}]$ is the shortest contiguous segment satisfying:
\begin{equation}
\begin{aligned}
N_{t_s^{(k)}-1} &= N_{t_e^{(k)}} ;=: N_{\mathrm{ref}},\
N_t &< N_{\mathrm{ref}}, \quad \forall, t \in [t_s^{(k)}, t_e^{(k)}-1].
\end{aligned}
\end{equation}
In other words, once the ID count drops below the reference value $N_{\mathrm{ref}}$, all frames until it returns to that value (or exceeds it) are regarded as belonging to a single occlusion event.
Let define the first-order difference as
\begin{equation}
\Delta N_t := N_t - N_{t-1}, \qquad t \ge 2,
\end{equation}
where $\Delta N_t < 0$ indicates a decrease in IDs and $\Delta N_t > 0$ indicates an increase.
The overall procedure of Occlusion-Session-Extractor is summarized in Algorithm~\ref{alg:occlusion}.


\begin{algorithm}[t]
\caption{OcclusionIntervalExtractor:}
\label{alg:occlusion}
\KwIn{track sets ${\mathcal{T}t} {t=1}^{T}$}
\KwOut{intervals ${I_k}$; lost IDs ${\tau^{(k)}{\text{lost}}}$; gained IDs ${\tau^{(k)}{\text{gain}}}$}
\DontPrintSemicolon
$t \gets 2$;
state $\gets$ \textsc{IDLE};
\While{$t \le T$}{
$N_t \gets |\mathcal{T}t|$;
\If{$state = \textsc{IDLE}$ \textbf{and} $N_t < N{t-1}$}{
state $\gets$ \textsc{OCCLUDE};
$t_s \gets t$;
$N_{\text{ref}} \gets N_{t-1}$;
$\tau_{\text{lost}} \gets \mathcal{T}{t-1} \setminus \mathcal{T}t$;
}
\ElseIf{$state = \textsc{OCCLUDE}$ \textbf{and} $N_t \ge N{\text{ref}}$}{
$t_e \gets t$;
$\tau{\text{gain}} \gets \mathcal{T}t \setminus \mathcal{T}{t-1}$;
\textsc{SaveInterval}$(t_s, t_e, \tau_{\text{lost}}, \tau_{\text{gain}})$;
state $\gets$ \textsc{IDLE};
}
$t \gets t + 1$;
}
\end{algorithm}

\begin{figure}[htbp]
  \centering
  \includegraphics[width=\columnwidth]{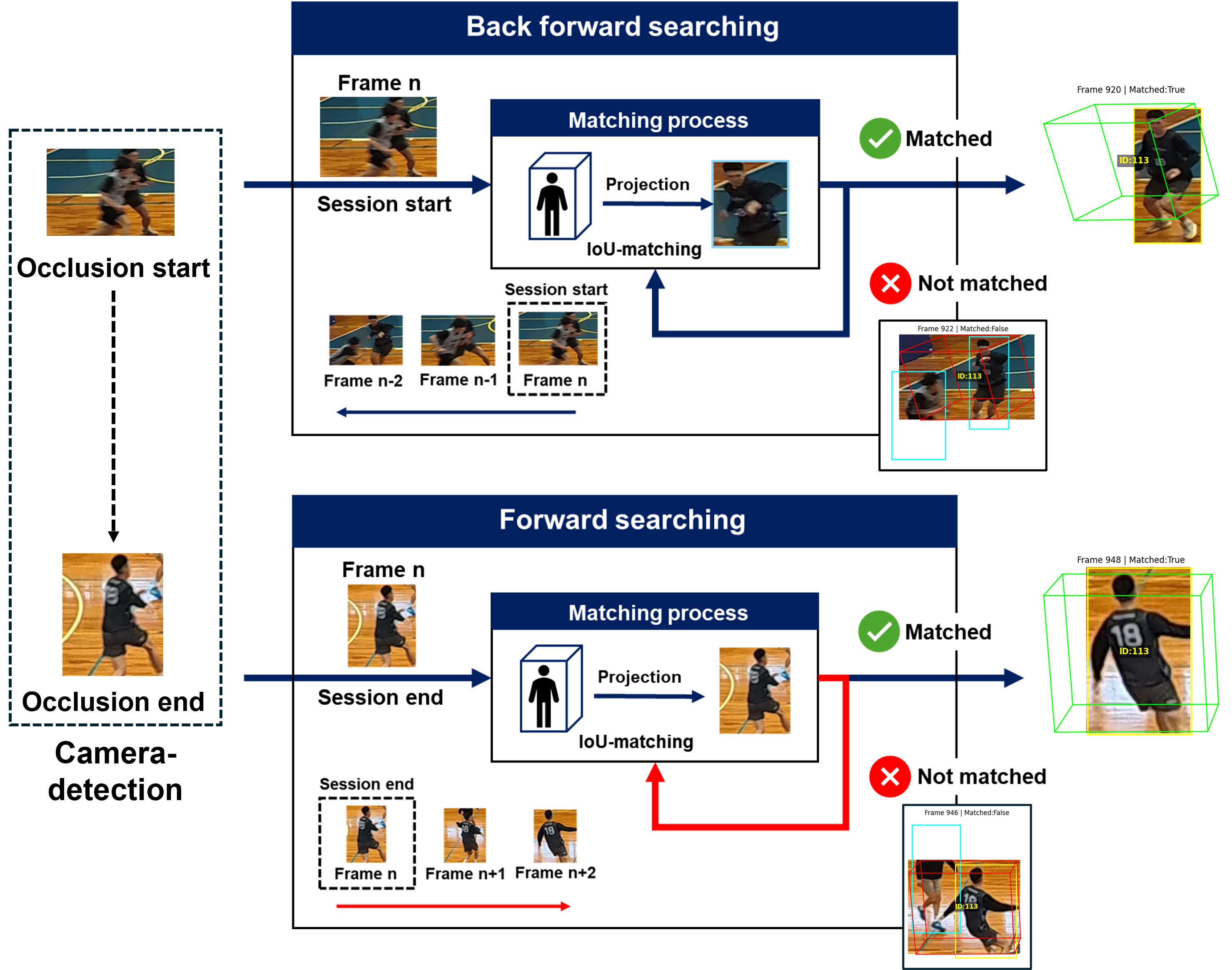}
  \caption{LiDAR-camera-detection-matching: Frames are scanned from the occlusion boundary in the selected camera. The target's 3D voxel is intersected with current image detections; the first frame with a unique overlap yields the clean reference patch for re-identification.}
  \label{…}
\end{figure}

\subsubsection{LiDAR-camera-detection-matching}
To improve the accuracy of the subsequent re-identification stage, an image patch in which the target ID is captured most clearly is extracted, starting from either the start frame $t_s^{(k)}$ or the end frame $t_e^{(k)}$ of occlusion interval $k$. Beginning at $t_s^{(k)}$ for a backward search and $t_e^{(k)}$ for a forward search, the \textit{Frame~Searching} procedure is applied recursively until a clear frame is obtained.

\noindent\textbf{Frame Searching.}
At frame $t$, every bounding box on BEV is projected onto all cameras, and the clarity of the corresponding player image before or after occlusion is evaluated. The entire procedure is summarised in Algorithm~\ref{alg:frameSelect}.

First, the bounding boxes on BEV $\{\tilde B_i(t)\}_{i=1}^{N}$ are transformed to the world coordinate system, given a height range $[z_{\min}, z_{\max}]$, to form three-dimensional voxels $\hat B_i(t)\subset\mathbb{R}^3$. Each voxel is then perspective-projected by the camera matrix $\Pi_c$ as
\begin{equation}
    B_i^{c}(t) = \Pi_c\!\bigl(\hat B_i(t)\bigr).
\end{equation}

Next, for each camera $c$ and its projected box $B_i^{c}(t)$, the number of corner points that fall inside the projection sets of other IDs, $\{B_j^{c}(t)\}_{j\neq i}$, is counted. The camera containing the fewest such points is regarded as capturing the occluded player most distinctly. Using the inverse of the projected box area as the score, the camera with the minimum score is selected:
\begin{equation}
    c^{\ast} = \arg\min_{c}\bigl(\text{area}(B_i^{c}(t))^{-1}\bigr).
\end{equation}

Then, YOLOv11~\cite{yolov11} is applied to the selected camera frame to obtain detections $\mathcal{B}_{\mathrm{YOLO}}$. The intersection over union between the projected box $B_{\mathrm{proj}}$ and $\mathcal{B}_{\mathrm{YOLO}}$ is compared. If exactly one detection overlaps $B_{\mathrm{proj}}$, the frame is regarded as clear; otherwise, if multiple detections overlap, the frame is deemed unclear, and the search continues.



\begin{algorithm}[t]
\DontPrintSemicolon
\SetKwInOut{Input}{Input}
\SetKwInOut{Output}{Output}
\Input{ID $i$, seed frame $t_0$, direction $dir\in\{\textit{backward},\textit{forward}\}$;\
       BEV boxes $\{\tilde B_k(t)\}_{k=1}^N$;\
       camera matrices $\{\Pi_c\}_{c=1}^C$}
\Output{Frame $F=(c^\ast,t)$ (\texttt{nil} if not found)}
$t\gets t_0$\;
\While{true}{
  $c^\ast\gets\texttt{nil}$;\quad $minInc\gets8$;\quad $bestScore\gets\infty$\;
  \For{$c\gets1$ \KwTo $C$}{
    $B_i\gets\Pi_c\bigl(\mathrm{Voxelize}(\tilde B_i(t))\bigr)$\;
    $inc\gets0$\;                                        
    \For{$p\in\mathrm{corners}(B_i)$}{
      \ForEach{$j\neq i$}{
        $B_j\gets\Pi_c\bigl(\mathrm{Voxelize}(\tilde B_j(t))\bigr)$\;
        \If{$p\in B_j$}{$inc\gets inc+1$;\quad\textbf{break}}
      }
    }
    $score\gets\textproc{DepthMetric}(i,c,t)$            
    \If{$inc<minInc$\ \textbf{or}\ ($inc=minInc$\ \textbf{and}\ $score<bestScore$)}{
        $minInc\gets inc$;\quad $bestScore\gets score$;\quad $c^\ast\gets c$
    }
  }
  \If{$c^\ast=\texttt{nil}$}{
    $t\gets t-1$ \textbf{if} $dir=\textit{backward}$ \textbf{else} $t\gets t+1$;\quad\textbf{continue}
  }
  $B_{\text{proj}}\gets\Pi_{c^\ast}\bigl(\mathrm{Voxelize}(\tilde B_i(t))\bigr)$\;
  $B_{\text{YOLO}}\gets\textproc{YOLO}(\textit{frame}(c^\ast,t))$\;
  $\mathcal{B}_{\text{high}}\gets\bigl\{\,b\in B_{\text{YOLO}}\mid
      \mathrm{IoU}(b,B_{\text{proj}})\ge\tau_{\text{high}}\bigr\}$\;

    
  \If{$|\mathcal{B}_{\text{high}}|=1$
      \textbf{and}
      $\displaystyle\max_{b\in B_{\text{YOLO}}\setminus\mathcal{B}_{\text{high}}}
       \mathrm{IoU}(b,B_{\text{proj}})<\tau_{\text{low}}$}{
       \Return{$(c^\ast,t)$}
  }
  $t\gets t-1$ \textbf{if} $dir=\textit{backward}$ \textbf{else} $t\gets t+1$
}
\caption{Camera search and frame selection}
\label{alg:frameSelect}
\end{algorithm}







\begin{table*}[t]
\centering
\caption{Overall comparison of MOT performance}
\label{tab:overall_performance}
\begin{tabular}{lcccccc}
\toprule
& HOTA$\uparrow$ & IDF1$\uparrow$ & AssA$\uparrow$ & MOTA$\uparrow$ & DetA$\uparrow$ & FPS$\uparrow$ \\
\midrule
LiDAR-camera-fusion & 0.917 & 0.930 & 0.881 & 0.957 & 0.955 & 6.34 \\
LiDAR-only       & 0.917 & 0.918 & 0.877 & 0.957 & 0.955 &  28.4 \\
camera-only         & 0.831 & 0.868 & 0.810 & 0.842 & 0.853 & 0.218 \\
\bottomrule
\end{tabular}
\end{table*}

\subsubsection{re-identification}
For each ID
$\tau_k^{\mathrm{lost}}, \tau_k^{\mathrm{near\text{-}lost}}, \tau_k^{\mathrm{gain}}, \tau_k^{\mathrm{near\text{-}gain}}$
involved in occlusion $k$ obtained by the \textit{Occlusion-Session-Extractor}, applying \textit{LiDAR–Camera–Detection-Matching} yields image patches in which both players are captured sharply before and after the occlusion, denoted as
$F_k^{\mathrm{lost}}, F_k^{\mathrm{near\text{-}lost}},  F_k^{\mathrm{gain}}, F_k^{\mathrm{near\text{-}gain}}$. 
Features are extracted from these patches with a ResNet-50~\cite{he2016deep} pre-trained on the Market1501 dataset~\cite{masson2021exploiting}, and cosine similarity is computed between the pre- and post-occlusion features.

\begin{equation}
    \mathbf{f}_k^{\bullet}=\phi\!\bigl(F_k^{\bullet}\bigr)
\in\mathbb{R}^{d},
\quad
\bullet\in\{
\mathrm{lost},\mathrm{near\text{-}lost},
\mathrm{gain},\mathrm{near\text{-}gain}\}.
\end{equation}


Finally, the pair with the highest cosine similarity is treated as the same player, and the post-occlusion ID is replaced with the pre-occlusion ID.


\section{Experiment}
In this study, we used the newly constructed BasketLiDAR dataset to validate the effectiveness of our proposed method by conducting the following experiments: (1) a comparative evaluation against the conventional camera-only approach; and (2) a comparative evaluation between LiDAR-only and LiDAR–camera fusion MOT pipelines. In each experiment, we assess multi-object tracking (MOT) performance in terms of both accuracy and processing speed, with the aim of quantitatively demonstrating the superiority of the proposed method.


\subsection{Experiment Setup}
We split the 34 recorded video clips into 24 training clips and 10 test clips.The data split was performed so that the difficulty of each clip (measured by the number of players and the frequency of occlusions) was balanced across the training and test sets. 



For a fair comparison, we employed YOLOv11 \cite{yolov11} as the common detector across all methods and retrained it on each method’s respective training split, using the default configuration and fine-tuning from the public weights. For tracking, we adopted ByteTrack \cite{zhang2022bytetrack} in our LiDAR-based pipeline and Observation-Centric SORT \cite{cao2023observation} in the camera-only pipeline. Since the camera-based pipeline rely on keypoint-based tracking, we employed OC-SORT for the tracker. Since the LiDAR-based pipeline can incorporate bbox-based tracking, we adopted ByteTrack due to its real-time tracking capability. Experiments were run on an AWS EC2 g4dn.xlarge instance, where we measured the inference throughput.



\subsection{Metrics}
For the evaluation of MOT performance, we used the following standard metrics: \textbf{MOTA}, which is computed from the counts of false positives (FP), false negatives (FN), and identity switches (IDSW) and primarily emphasizes detection performance; \textbf{IDF1}, which evaluates ID association performance; and \textbf{HOTA}, which provides a unified assessment of both detection and association performance. The HOTA metric can be further decomposed into the \textbf{DetA} and \textbf{AssA} submetrics, focusing respectively on detection and association.

To assess recovery performance from occlusions, we defined the ID recovery rate \(R_{\mathrm{ID}}\) as the frequency with which a ground-truth ID and tracker ID pair that was once lost is subsequently tracked as the same pair. This rate is expressed as the ratio of the number of re-matched pairs \(N_{\mathrm{re}}\) to the number of ID update or disappearance events \(N_{\mathrm{dis}}\) during tracking, i.e.,
\begin{equation}
    R_{\mathrm{ID}} = \frac{N_{\mathrm{re}}}{N_{\mathrm{dis}}}\,.
\end{equation}

Since the annotations are 3D, all evaluations were conducted as point-based metrics, in which similarity is calculated using distances between detection points instead of an IoU threshold. The distance threshold between detection points was set to the mean diagonal length of the ground-truth bounding boxes.





\begin{table}[t]
  \centering
  \caption{Inference speed (ms/frame)}
  \label{tab:processing_time}
  \resizebox{\columnwidth}{!}{%
    \begin{tabular}{lccc} 
      \toprule
        & Camera-only       & LiDAR-only     & LiDAR-camera-fusion \\
      \midrule
      Detection \& Tracking   & 628.0   & 35.2  & 35.2 \\
      Triangulation        & 3954.5 & --     & -- \\
      LiDAR-camera-fusion ReID      & --     & --    & 122.6 \\
      Total                & 4582.5 & 35.2  & 157.8 \\
      \bottomrule
    \end{tabular}%
  }
\end{table}

\begin{table}[t]
  \centering                      
  \caption{Comparison between LiDAR-only and LiDAR-camera-fusion}
  \label{tab:comparison-fusion}
  \footnotesize                   
  \begin{tabular}{l c}            
    \toprule
    Method & \begin{tabular}{c} Occlusion \\Recovery Rate$\uparrow$ \end{tabular} \\
    \midrule
    LiDAR-camera-fusion & 0.241 \\
    LiDAR-only          & 0.158 \\
    \bottomrule
  \end{tabular}
\end{table}


\subsection{Results}
\subsubsection{Camera-only Vs.\ LiDAR-camera-fusion.}
Table \ref{tab:overall_performance} compares the overall performance of the proposed LiDAR–camera-fusion, LiDAR-only, and camera-only methods. The proposed approach achieves substantial improvements over the camera-only baseline across all metrics. Notably, it attains improves of 11.5\% in MOTA and 10.2\% in DetA, metrics that emphasize detection accuracy. We attribute these gains to the explicit 3D spatial information provided by LiDAR, which leads to a marked increase in true-positive detections.

The improvement in processing speed is significant, as shown in Table~\ref{tab:processing_time}. The LiDAR-only method achieves approximately a $130\times$ speedup (0.22 FPS → 28.4 FPS), and the LiDAR–camera fusion method achieves approximately a $30\times$ speedup (0.22 FPS → 6.34 FPS) compared to the camera-only approach. The primary reason for this dramatic speed-up  is that the triangulation processing—which accounted for 86\% of the total processing time in the camera-only method (3954.5 ms/frame)—becomes unnecessary thanks to LiDAR’s direct acquisition of 3D information.


\subsubsection{LiDAR-only Vs.\  LiDAR-camera-fusion}
Table~\ref{tab:comparison-fusion} presents the quantitative evaluation using ID-related metrics, showing that camera fusion improved the ID recovery rate from 0.16 to 0.24, confirming enhanced re-identification performance after extended occlusions. This improvement is attributed to the integration of LiDAR positional information with camera appearance features, enabling accurate re-identification post-occlusion. Regarding processing speed, although the introduction of camera fusion leads to a decrease (28.4 FPS → 6.3 FPS), it still maintains sufficient performance for real-time processing.





\section{Conclusion}
In this study, we proposed a novel approach combining LiDAR and cameras to overcome the limitations of conventional camera-only methods in the sports 3D MOT field. We constructed BasketLiDAR, the world's first multimodal MOT dataset for sports, providing professional-level data from actual basketball players. The proposed LiDAR-camera-fusion framework achieved improved accuracy and significant processing speed enhancement compared to camera-based methods, while also improving ID re-identification performance under occlusion conditions. Through direct 3D information acquisition from LiDAR, we realized high-precision and real-time sports MOT. The BasketLiDAR dataset will be open-sourced, and we expect it to contribute to research advancement in the sports MOT field and the proliferation of data-driven sports analysis.

\section*{Acknowledgements}
This research was supported in part by the  JST CREST JPMJCR23M4, JST PRESTO JPMJPR22PA, and JSPS KAKENHI 24K02940.

\bibliographystyle{ACM-Reference-Format}
\bibliography{citations}


\begin{thebibliography}{23}


\ifx \showCODEN    \undefined \def \showCODEN     #1{\unskip}     \fi
\ifx \showISBNx    \undefined \def \showISBNx     #1{\unskip}     \fi
\ifx \showISBNxiii \undefined \def \showISBNxiii  #1{\unskip}     \fi
\ifx \showISSN     \undefined \def \showISSN      #1{\unskip}     \fi
\ifx \showLCCN     \undefined \def \showLCCN      #1{\unskip}     \fi
\ifx \shownote     \undefined \def \shownote      #1{#1}          \fi
\ifx \showarticletitle \undefined \def \showarticletitle #1{#1}   \fi
\ifx \showURL      \undefined \def \showURL       {\relax}        \fi
\providecommand\bibfield[2]{#2}
\providecommand\bibinfo[2]{#2}
\providecommand\natexlab[1]{#1}
\providecommand\showeprint[2][]{arXiv:#2}

\bibitem[ins({[n.\,d.]})]%
        {instaace}
 \bibinfo{year}{[n.\,d.]}\natexlab{}.
\newblock \bibinfo{title}{{Insta360 Ace Pro}}.
\newblock \bibinfo{howpublished}{\url{https://www.insta360.com/product/insta360-ace-pro}}.
\newblock


\bibitem[liv({[n.\,d.]})]%
        {livox-hap}
 \bibinfo{year}{[n.\,d.]}\natexlab{}.
\newblock \bibinfo{title}{{Livox HAP LiDAR}}.
\newblock \bibinfo{howpublished}{\url{https://www.livoxtech.com/hap}}.
\newblock


\bibitem[Bewley et~al\mbox{.}(2016)]%
        {bewley2016simple}
\bibfield{author}{\bibinfo{person}{Alex Bewley}, \bibinfo{person}{Zongyuan Ge}, \bibinfo{person}{Lionel Ott}, \bibinfo{person}{Fabio Ramos}, {and} \bibinfo{person}{Ben Upcroft}.} \bibinfo{year}{2016}\natexlab{}.
\newblock \showarticletitle{Simple online and realtime tracking}. In \bibinfo{booktitle}{\emph{2016 IEEE international conference on image processing (ICIP)}}. Ieee, \bibinfo{pages}{3464--3468}.
\newblock


\bibitem[Caesar et~al\mbox{.}(2020)]%
        {caesar2020nuscenes}
\bibfield{author}{\bibinfo{person}{Holger Caesar}, \bibinfo{person}{Varun Bankiti}, \bibinfo{person}{Alex~H Lang}, \bibinfo{person}{Sourabh Vora}, \bibinfo{person}{Venice~Erin Liong}, \bibinfo{person}{Qiang Xu}, \bibinfo{person}{Anush Krishnan}, \bibinfo{person}{Yu Pan}, \bibinfo{person}{Giancarlo Baldan}, {and} \bibinfo{person}{Oscar Beijbom}.} \bibinfo{year}{2020}\natexlab{}.
\newblock \showarticletitle{nuscenes: A multimodal dataset for autonomous driving}. In \bibinfo{booktitle}{\emph{Proceedings of the IEEE/CVF conference on computer vision and pattern recognition}}. \bibinfo{pages}{11621--11631}.
\newblock


\bibitem[Cao et~al\mbox{.}(2023)]%
        {cao2023observation}
\bibfield{author}{\bibinfo{person}{Jinkun Cao}, \bibinfo{person}{Jiangmiao Pang}, \bibinfo{person}{Xinshuo Weng}, \bibinfo{person}{Rawal Khirodkar}, {and} \bibinfo{person}{Kris Kitani}.} \bibinfo{year}{2023}\natexlab{}.
\newblock \showarticletitle{Observation-centric sort: Rethinking sort for robust multi-object tracking}. In \bibinfo{booktitle}{\emph{Proceedings of the IEEE/CVF conference on computer vision and pattern recognition}}. \bibinfo{pages}{9686--9696}.
\newblock


\bibitem[Chen et~al\mbox{.}(2017)]%
        {chen2017multi}
\bibfield{author}{\bibinfo{person}{Xiaozhi Chen}, \bibinfo{person}{Huimin Ma}, \bibinfo{person}{Ji Wan}, \bibinfo{person}{Bo Li}, {and} \bibinfo{person}{Tian Xia}.} \bibinfo{year}{2017}\natexlab{}.
\newblock \showarticletitle{Multi-view 3d object detection network for autonomous driving}. In \bibinfo{booktitle}{\emph{Proceedings of the IEEE conference on Computer Vision and Pattern Recognition}}. \bibinfo{pages}{1907--1915}.
\newblock


\bibitem[Cui et~al\mbox{.}(2023)]%
        {cui2023sportsmot}
\bibfield{author}{\bibinfo{person}{Yutao Cui}, \bibinfo{person}{Chenkai Zeng}, \bibinfo{person}{Xiaoyu Zhao}, \bibinfo{person}{Yichun Yang}, \bibinfo{person}{Gangshan Wu}, {and} \bibinfo{person}{Limin Wang}.} \bibinfo{year}{2023}\natexlab{}.
\newblock \showarticletitle{Sportsmot: A large multi-object tracking dataset in multiple sports scenes}. In \bibinfo{booktitle}{\emph{Proceedings of the IEEE/CVF international conference on computer vision}}. \bibinfo{pages}{9921--9931}.
\newblock


\bibitem[Dong et~al\mbox{.}(2019)]%
        {dong2019fast}
\bibfield{author}{\bibinfo{person}{Junting Dong}, \bibinfo{person}{Wen Jiang}, \bibinfo{person}{Qixing Huang}, \bibinfo{person}{Hujun Bao}, {and} \bibinfo{person}{Xiaowei Zhou}.} \bibinfo{year}{2019}\natexlab{}.
\newblock \showarticletitle{Fast and robust multi-person 3d pose estimation from multiple views}. In \bibinfo{booktitle}{\emph{Proceedings of the IEEE/CVF conference on computer vision and pattern recognition}}. \bibinfo{pages}{7792--7801}.
\newblock


\bibitem[Geiger et~al\mbox{.}(2012)]%
        {geiger2012kitti}
\bibfield{author}{\bibinfo{person}{Andreas Geiger}, \bibinfo{person}{Philip Lenz}, {and} \bibinfo{person}{Raquel Urtasun}.} \bibinfo{year}{2012}\natexlab{}.
\newblock \showarticletitle{Are we ready for autonomous driving? the kitti vision benchmark suite}. In \bibinfo{booktitle}{\emph{2012 IEEE conference on computer vision and pattern recognition}}. IEEE, \bibinfo{pages}{3354--3361}.
\newblock


\bibitem[Han et~al\mbox{.}(2019)]%
        {han2019re}
\bibfield{author}{\bibinfo{person}{Chuchu Han}, \bibinfo{person}{Jiacheng Ye}, \bibinfo{person}{Yunshan Zhong}, \bibinfo{person}{Xin Tan}, \bibinfo{person}{Chi Zhang}, \bibinfo{person}{Changxin Gao}, {and} \bibinfo{person}{Nong Sang}.} \bibinfo{year}{2019}\natexlab{}.
\newblock \showarticletitle{Re-id driven localization refinement for person search}. In \bibinfo{booktitle}{\emph{Proceedings of the IEEE/CVF International Conference on Computer Vision}}. \bibinfo{pages}{9814--9823}.
\newblock


\bibitem[He et~al\mbox{.}(2016)]%
        {he2016deep}
\bibfield{author}{\bibinfo{person}{Kaiming He}, \bibinfo{person}{Xiangyu Zhang}, \bibinfo{person}{Shaoqing Ren}, {and} \bibinfo{person}{Jian Sun}.} \bibinfo{year}{2016}\natexlab{}.
\newblock \showarticletitle{Deep residual learning for image recognition}. In \bibinfo{booktitle}{\emph{Proceedings of the IEEE conference on computer vision and pattern recognition}}. \bibinfo{pages}{770--778}.
\newblock


\bibitem[He et~al\mbox{.}(2023)]%
        {he2023fastreid}
\bibfield{author}{\bibinfo{person}{Lingxiao He}, \bibinfo{person}{Xingyu Liao}, \bibinfo{person}{Wu Liu}, \bibinfo{person}{Xinchen Liu}, \bibinfo{person}{Peng Cheng}, {and} \bibinfo{person}{Tao Mei}.} \bibinfo{year}{2023}\natexlab{}.
\newblock \showarticletitle{Fastreid: A pytorch toolbox for general instance re-identification}. In \bibinfo{booktitle}{\emph{Proceedings of the 31st ACM International Conference on Multimedia}}. \bibinfo{pages}{9664--9667}.
\newblock


\bibitem[Jocher et~al\mbox{.}(2023)]%
        {yolov11}
\bibfield{author}{\bibinfo{person}{Glenn Jocher}, \bibinfo{person}{Jing Qiu}, {and} \bibinfo{person}{Ayush Chaurasia}.} \bibinfo{year}{2023}\natexlab{}.
\newblock \bibinfo{booktitle}{\emph{{Ultralytics YOLOv11}}}.
\newblock
\urldef\tempurl%
\url{https://github.com/ultralytics/ultralytics}
\showURL{%
\tempurl}


\bibitem[Lang et~al\mbox{.}(2019)]%
        {lang2019pointpillars}
\bibfield{author}{\bibinfo{person}{Alex~H Lang}, \bibinfo{person}{Sourabh Vora}, \bibinfo{person}{Holger Caesar}, \bibinfo{person}{Lubing Zhou}, \bibinfo{person}{Jiong Yang}, {and} \bibinfo{person}{Oscar Beijbom}.} \bibinfo{year}{2019}\natexlab{}.
\newblock \showarticletitle{Pointpillars: Fast encoders for object detection from point clouds}. In \bibinfo{booktitle}{\emph{Proceedings of the IEEE/CVF conference on computer vision and pattern recognition}}. \bibinfo{pages}{12697--12705}.
\newblock


\bibitem[Masson et~al\mbox{.}(2021)]%
        {masson2021exploiting}
\bibfield{author}{\bibinfo{person}{Hugo Masson}, \bibinfo{person}{Amran Bhuiyan}, \bibinfo{person}{Le~Thanh Nguyen-Meidine}, \bibinfo{person}{Mehrsan Javan}, \bibinfo{person}{Parthipan Siva}, \bibinfo{person}{Ismail~Ben Ayed}, {and} \bibinfo{person}{Eric Granger}.} \bibinfo{year}{2021}\natexlab{}.
\newblock \showarticletitle{Exploiting prunability for person re-identification}.
\newblock \bibinfo{journal}{\emph{EURASIP Journal on Image and Video Processing}} \bibinfo{volume}{2021}, \bibinfo{number}{1} (\bibinfo{year}{2021}), \bibinfo{pages}{22}.
\newblock


\bibitem[Qi et~al\mbox{.}(2018)]%
        {qi2018frustum}
\bibfield{author}{\bibinfo{person}{Charles~R Qi}, \bibinfo{person}{Wei Liu}, \bibinfo{person}{Chenxia Wu}, \bibinfo{person}{Hao Su}, {and} \bibinfo{person}{Leonidas~J Guibas}.} \bibinfo{year}{2018}\natexlab{}.
\newblock \showarticletitle{Frustum pointnets for 3d object detection from rgb-d data}. In \bibinfo{booktitle}{\emph{Proceedings of the IEEE conference on computer vision and pattern recognition}}. \bibinfo{pages}{918--927}.
\newblock


\bibitem[Scott et~al\mbox{.}(2024)]%
        {scott2024teamtrack}
\bibfield{author}{\bibinfo{person}{Atom Scott}, \bibinfo{person}{Ikuma Uchida}, \bibinfo{person}{Ning Ding}, \bibinfo{person}{Rikuhei Umemoto}, \bibinfo{person}{Rory Bunker}, \bibinfo{person}{Ren Kobayashi}, \bibinfo{person}{Takeshi Koyama}, \bibinfo{person}{Masaki Onishi}, \bibinfo{person}{Yoshinari Kameda}, {and} \bibinfo{person}{Keisuke Fujii}.} \bibinfo{year}{2024}\natexlab{}.
\newblock \showarticletitle{Teamtrack: A dataset for multi-sport multi-object tracking in full-pitch videos}. In \bibinfo{booktitle}{\emph{Proceedings of the IEEE/CVF conference on computer vision and pattern recognition}}. \bibinfo{pages}{3357--3366}.
\newblock


\bibitem[Sun et~al\mbox{.}(2020)]%
        {sun2020waymo}
\bibfield{author}{\bibinfo{person}{Pei Sun}, \bibinfo{person}{Henrik Kretzschmar}, \bibinfo{person}{Xerxes Dotiwalla}, \bibinfo{person}{Aurelien Chouard}, \bibinfo{person}{Vijaysai Patnaik}, \bibinfo{person}{Paul Tsui}, \bibinfo{person}{James Guo}, \bibinfo{person}{Yin Zhou}, \bibinfo{person}{Yuning Chai}, \bibinfo{person}{Benjamin Caine}, {et~al\mbox{.}}} \bibinfo{year}{2020}\natexlab{}.
\newblock \showarticletitle{Scalability in perception for autonomous driving: Waymo open dataset}. In \bibinfo{booktitle}{\emph{Proceedings of the IEEE/CVF conference on computer vision and pattern recognition}}. \bibinfo{pages}{2446--2454}.
\newblock


\bibitem[Weng et~al\mbox{.}(2020)]%
        {weng2020ab3dmot}
\bibfield{author}{\bibinfo{person}{Xinshuo Weng}, \bibinfo{person}{Jianren Wang}, \bibinfo{person}{David Held}, {and} \bibinfo{person}{Kris Kitani}.} \bibinfo{year}{2020}\natexlab{}.
\newblock \showarticletitle{Ab3dmot: A baseline for 3d multi-object tracking and new evaluation metrics}.
\newblock \bibinfo{journal}{\emph{arXiv preprint arXiv:2008.08063}} (\bibinfo{year}{2020}).
\newblock


\bibitem[Wojke et~al\mbox{.}(2017)]%
        {wojke2017simple}
\bibfield{author}{\bibinfo{person}{Nicolai Wojke}, \bibinfo{person}{Alex Bewley}, {and} \bibinfo{person}{Dietrich Paulus}.} \bibinfo{year}{2017}\natexlab{}.
\newblock \showarticletitle{Simple online and realtime tracking with a deep association metric}. In \bibinfo{booktitle}{\emph{2017 IEEE international conference on image processing (ICIP)}}. IEEE, \bibinfo{pages}{3645--3649}.
\newblock


\bibitem[Wu et~al\mbox{.}(2020)]%
        {wu2020multi}
\bibfield{author}{\bibinfo{person}{Wanneng Wu}, \bibinfo{person}{Min Xu}, \bibinfo{person}{Qiaokang Liang}, \bibinfo{person}{Li Mei}, {and} \bibinfo{person}{Yu Peng}.} \bibinfo{year}{2020}\natexlab{}.
\newblock \showarticletitle{Multi-camera 3D ball tracking framework for sports video}.
\newblock \bibinfo{journal}{\emph{IET Image Processing}} \bibinfo{volume}{14}, \bibinfo{number}{15} (\bibinfo{year}{2020}), \bibinfo{pages}{3751--3761}.
\newblock


\bibitem[Yamada et~al\mbox{.}(2025)]%
        {yamada2025trackid3x3}
\bibfield{author}{\bibinfo{person}{Kazuhiro Yamada}, \bibinfo{person}{Li Yin}, \bibinfo{person}{Qingrui Hu}, \bibinfo{person}{Ning Ding}, \bibinfo{person}{Shunsuke Iwashita}, \bibinfo{person}{Jun Ichikawa}, \bibinfo{person}{Kiwamu Kotani}, \bibinfo{person}{Calvin Yeung}, {and} \bibinfo{person}{Keisuke Fujii}.} \bibinfo{year}{2025}\natexlab{}.
\newblock \showarticletitle{TrackID3x3: A Dataset and Algorithm for Multi-Player Tracking with Identification and Pose Estimation in 3x3 Basketball Full-court Videos}.
\newblock \bibinfo{journal}{\emph{arXiv preprint arXiv:2503.18282}} (\bibinfo{year}{2025}).
\newblock


\bibitem[Zhang et~al\mbox{.}(2022)]%
        {zhang2022bytetrack}
\bibfield{author}{\bibinfo{person}{Yifu Zhang}, \bibinfo{person}{Peize Sun}, \bibinfo{person}{Yi Jiang}, \bibinfo{person}{Dongdong Yu}, \bibinfo{person}{Fucheng Weng}, \bibinfo{person}{Zehuan Yuan}, \bibinfo{person}{Ping Luo}, \bibinfo{person}{Wenyu Liu}, {and} \bibinfo{person}{Xinggang Wang}.} \bibinfo{year}{2022}\natexlab{}.
\newblock \showarticletitle{ByteTrack: Multi-Object Tracking by Associating Every Detection Box}. In \bibinfo{booktitle}{\emph{Proceedings of the European Conference on Computer Vision (ECCV)}}. Springer, \bibinfo{pages}{1--21}.
\newblock


\end{thebibliography}

\end{document}